# Prediction of the progression of subcortical brain structures in Alzheimer's disease from baseline

Alexandre Bône[†12], Maxime Louis[†12], Alexandre Routier[12], Jorge Samper[12], Michael Bacci[12], Benjamin Charlier[23], Olivier Colliot[12], Stanley Durrleman[12], and the Alzheimer's Disease Neuroimaging Initiative

[1] Sorbonne Universités, UPMC Université Paris 06, Inserm, CNRS, Institut du Cerveau et de la Moelle (ICM) – Hôpital Pitié-Salpêtrière, 75013 Paris, France,
[2] Inria Paris, Aramis project-team, 75013 Paris, France,
[3] Université de Montpellier, France

**Abstract.** We propose a method to predict the subject-specific longitudinal progression of brain structures extracted from baseline MRI, and evaluate its performance on Alzheimer's disease data. The disease progression is modeled as a trajectory on a group of diffeomorphisms in the context of large deformation diffeomorphic metric mapping (LDDMM). We first exhibit the limited predictive abilities of geodesic regression extrapolation on this group. Building on the recent concept of parallel curves in shape manifolds, we then introduce a second predictive protocol which personalizes previously learned trajectories to new subjects, and investigate the relative performances of two parallel shifting paradigms. This design only requires the baseline imaging data. Finally, coefficients encoding the disease dynamics are obtained from longitudinal cognitive measurements for each subject, and exploited to refine our methodology which is demonstrated to successfully predict the follow-up visits.

## 1 Introduction

The primary pathological developments of a neurodegenerative disease such as Alzheimer's are believed to spring long before the first symptoms of cognitive decline. Subtle gradual structural alterations of the brain arise and develop along the disease course, in particular in the hippocampi regions, whose volumes are classical biomarkers in clinical trials. Among other factors, those transformations ultimately result in the decline of cognitive functions, which can be assessed through standardized tests. Being able to track and predict future structural changes in the brain is therefore key to estimate the individual stage of disease progression, to select patients and provide endpoints in clinical trials.

To this end, our work settles down to predict the future shape of brain structures segmented from MRIs. We propose a methodology based on three building blocks : extrapolate from the past of a subject ; transfer the progression of a reference subject observed over a longer time period to new subjects ;

[†]Equal contributions.

and refine this transfer with information about the relative disease dynamics extracted from cognitive evaluations. Instead of limiting ourselves to specific features such as volumes, we propose to see each observation of a patient at a given time-point as an image or a segmented surface mesh in a shape space.

In computational anatomy, shape spaces are usually defined via the action of a group of diffeomorphisms [1,16,17]. In this framework, one may estimate a flow of diffeomorphisms such that a shape continuously deformed by this flow best fits repeated observations of the same subject over time, thus leading to a subject-specific spatiotemporal trajectory of shape changes [8,12]. If the flow is geodesic in the sense of a shortest path in the group of diffeomorphisms, this problem is called geodesic regression [4,5,8,12] and may be thought of as the extension to Riemannian manifolds of the linear regression concept. It is tempting then to use such regression to infer the future evolution of the shape given several past observations. To the best of our knowledge, the predictive power of such a method has not yet been extensively assessed. We will demonstrate that satisfying results can only be obtained when large numbers of data points over extensive periods of time are available, and that poor ones should be expected in the more interesting use-case scenario of a couple of observations.

In such situations, an appealing workaround would be to transfer previously acquired knowledge from another patient observed over a longer period of time. This idea requires the definition of a spatiotemporal matching method to transport the trajectory of shape changes into a different subject space. Several techniques have been proposed to register image time series of different subjects [11,18]. They often require time series to have the same number of images, or to have correspondences between images across time series, and are therefore unfit for prognosis purposes. Parallel transport in groups of diffeomorphisms has been recently introduced to infer deformation of follow-up images from baseline matching [10,15]. Such paradigms have been used mostly to transport spatiotemporal trajectories to the same anatomical space for hypothesis testing [6,13]. Two main methodologies have emerged: either by parallel-transporting the time series along the baseline matching as in [5], or by parallel-transporting the baseline matching along the time series as in [14]. We evaluate both in this paper.

In any case, these approaches require to match the baseline shape with one in the reference time series. Ideally, we should match observations corresponding to the same disease stage, which is unknown. We propose to complement such approaches with estimates of the patient stage and pace of progression using repeated neuropsychological assessments in the spirit of [14]. These estimates are used to adjust the dynamics of shape changes of the reference subject to the test one, according to the dynamical differences observed in the cognitive tests.

Among the main contributions of this papers are : the first quantitative study of the predictive power of geodesic regression ; a new methodology for the prediction of shape progression from baseline ; the evaluation of its accuracy for two different parallel shifting protocols ; new evidence of the utter importance of capturing the individual dynamics in Alzheimer's disease models.

Section 2 sets the theoretical background and incrementally describes our methodology. Section 3 presents and discusses the resulting performances.

## 2   Method

Let $(y_j)_{j=1,..,n_i}$ be a time series of segmented surface meshes for a given subject $i \in \{1, ..., N\}$, obtained at the ages $(t_j)_{j=1,...,n_i}$. We build a group of diffeomorphisms of the ambient space which act on the segmented meshes, following the procedure described in [3]. Flows of diffeomorphisms of $\mathbb{R}^3$ are generated by integrating time-varying vector fields of the form $v(t,x) = \sum_{k=1}^{n_{cp}} K[x, c_k(t)]\beta_k(t)$ where $K$ is a Gaussian kernel, $c(t) = [c_k(t)]_{k=1,..,n_{cp}}$ and $\beta(t) = [\beta_k(t)]_{k=1,..,n_{cp}}$ are respectively the control points and the momenta of the deformation.

We endow the space of diffeomorphisms with a norm which measures the cost of the deformation. In the following, we only consider geodesic flows of diffeomorphisms i.e. flows of minimal norm connecting the identity to a given diffeomorphism. Such flows are uniquely parametrized by their initial control points and momenta $c^0 = c(0)$, $\beta^0 = \beta(0)$. Under the action of the flow of diffeomorphisms, an initial template shape $T$ is continuously deformed and describes a trajectory in the shape space, which we will note $t \to \gamma_{(c^0,\beta^0)}(T,t)$. Simultaneously, we endow the surface meshes with a varifold norm $\|\cdot\|$ which allows to measure a data attachment term between meshes without point correspondence [3].

### 2.1   Geodesic regression

In the spirit of linear regression, one can perform geodesic regression in the shape space by estimating the intercept $T$ and the slope $(c^0, \beta^0)$ such that $\gamma_{(c^0,\beta^0)}(T, \cdot)$ minimizes the following functional :

$$\inf_{c^0,\beta^0,T} \sum_{j=1}^{n_i} \|\gamma_{(c^0,\beta^0)}(T,t_j) - y_j\|^2 + R(c^0, \beta^0) \qquad (1)$$

where $R$ is a regularization term which penalizes the kinetic energy of the deformation. We estimate a solution of equation (1) with a Nesterov gradient descent as implemented in the software Deformetrica (www.deformetrica.org), where the gradient with respect to the control points, the momenta and the template is computed with a backward integration of the data attachement term along the geodesic [2].

Once an optimum is found, we obtain a description of the progression of the brain structures which lies in the tangent space at the identity of the group of diffeomorphisms. It is natural to attempt to extrapolate from the obtained geodesic to obtain a prediction of the progression of the structures.

### 2.2   Two methods to transport spatiotemporal trajectories of shapes

As it will be demonstrated in section 3, geodesic regression extrapolation produces an accurate prediction only if data over a long time span is available for the subject, which is not compatible with the goal of early prognosis.

As proposed in [10,19], given a reference geodesic, we use the Riemannian parallel transport to generate a new trajectory. We first perform a baseline matching between the reference subject and the new subject, which can be described as a vector in the tangent space of the group of diffeomorphisms. Two paradigms are available to obtain a parallel trajectory. [15] advises to transport the reference regression along the matching and then shoot. In the shape space, this generates a geodesic starting at the baseline shape ; for this reason, we call this solution *geodesic parallelization*, and is illustrated on Figure (A1). On the other hand, [14] advocates to transport the matching vector along the reference geodesic and then build a trajectory with this transported vector from every point of the reference geodesic, as described on Figure (B1). We will call this procedure *exp-parallelization*.

In such a high-dimensional setting, the computation of parallel transport classically relies on the Schild's ladder scheme [9]. However, in our case the computation of the Riemannian logarithm may only be computed by solving a shape matching problem, resulting not only in an computationally expensive algorithm but also in an uncontrolled approximation of the scheme. To implement these parallel shifting methods, we use the algorithm suggested in [19], which relies on an approximation of the transport to nearby points by a well-chosen Jacobi field, with a sharp control on the computational complexity. The same rate of convergence as Schild's ladder is obtained at a reduced cost.

### 2.3 Cognitive scores dynamics

The protocol described in the previous section has two main drawbacks. First, the choice of the matching time in the reference trajectory is arbitrary : the baseline is purely a convenience choice and ideally the matching should be performed at similar stages of the disease. Second, it does not take into account the pace of progression of the subject. In [14], the authors propose a statistical model allowing to learn, in an unsupervised manner, dynamical parameters of the subjects from ADAS-cog test results, a standardized cognitive test designed for disease progression tracking. More specifically, they suppose that each patient follows a parallel to a mean trajectory, with a time reparametrization :

$$\psi(t) = \alpha(t - t_0 - \tau) + t_0 \qquad (2)$$

which maps the subject time to a normalized time frame, where $\alpha > 0$ and $\tau$ are scalar parameters. A high (resp. low) $\alpha$ hence corresponds to a fast (resp. slow) progression of the scores, when a negative (resp. positive) $\tau$ corresponds to an early decay (resp. late decay) of those scores. In the dataset introduced below, the acceleration factors $(\alpha_i)_i$ range from 0.15 to 6.01 and the time-shifts $(\tau_i)_i$ from $-20.6$ to $22.8$, thus showing a tremendous variability in the individual dynamics of the disease, which must be taken into account.

With these dynamic parameters, the shape evolution can be adjusted by reparametrizing the parallel trajectory with the same formula (2), as illustrated on Figures (A2) and (B2).

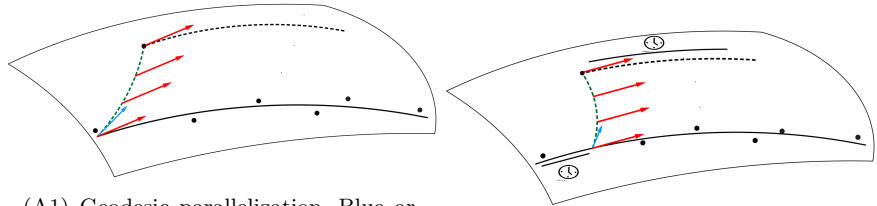

(A1) Geodesic parallelization. Blue arrow: baseline matching. Red arrows: transported regression. Black dotted line : exponentiation of the transported regression.

(A2) Reparametrized geodesic parallelization. Matching time and exp-parallel trajectory are reparametrized.

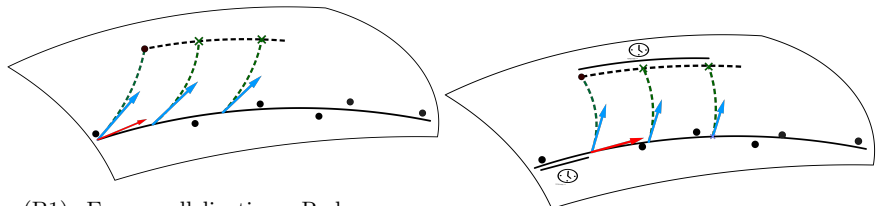

(B1) Exp-parallelization. Red arrow: geodesic regression. Blue arrows: transported baseline matching. Black dotted line : exp-parallelization of the reference geodesic for the given subject.

(B2) Reparametrized exp-parallelization. Matching time and exp-parallell trajectory are reparametrized.

## 3 Results

### 3.1 Data, preprocessing, parameters and performance metric

MRIs are extracted from the ADNI database, where only MCI converters with 7 visits or more are kept, for a total of N=74 subjects and 634 visits. Subjects are observed for a period of time ranging from 4 to 9 years (5.9 on average), with 12 visits at most. The 634 MRIs are segmented using the FreeSurfer software. The extracted brain masks are then affinely registered towards the Colin 27 Average Brain using the FSL software. The estimated transformations are finally applied to the pairs of caudates, hippocampi and putamina subcortical structures.

All diffeomorphic operations i.e. matching, geodesic regression estimation, shooting, exp-parallelization and geodesic parallelization are performed thanks to the Deformetrica software previously mentioned. A varifold distance with Gaussian kernel width of 3 mm for each structure and a deformation kernel width of 5 mm are chosen. The time discretization resolution is set to 2 months.

The chosen performance metric between two sets of meshes is the Dice coefficient, that is the sum of the volumes of the intersections of the corresponding meshes, divided by the total sum of the volumes. We only measure the volume of the intersection between corresponding structures. The Dice coefficient is comprised between 0 and 1 : it equals 1 for a perfect match, and 0 for disjoint structures.

## 3.2 Geodesic regression extrapolation

The acceleration factor $\alpha$ in equation (2) encodes the rate of progression of each patient. Multiplying this coefficient with the actual observation window gives a notion of the absolute observation window length, in the disease time referential. Only the 22 first subjects according to this measure have been considered for this section : they are indeed expected to feature large structural alterations, making the geodesic regression procedure more accurate. The geodesic regression predictive performance is compared to a naive one consisting in leaving the last observed brain structures in the learning dataset unchanged.

Table 1 presents the results obtained for varying learning dataset and extrapolation extents. We perform a Mann-Whitney test with the null hypothesis that the observed Dice coefficients distributions are the same to obtain the statistical significance levels. The extrapolated meshes are satisfying only in the case where all but one data points are used to perform the geodesic regression, achieving a high Dice index and outperforming the naive one, by a small margin though and failing to reach the significance level (p=0.25). When the window of observation becomes narrower, the prediction accuracy decreases and becomes worse than the naive one. Indeed, the lack of robustness of the – although standard – segmentation pipeline imposes a high noise level, which seems to translate into a too low signal-to-noise ratio after extrapolation from only a few observations.

| Learning period (months) | Method | Predicted follow-up visit | | | | | |
|---|---|---|---|---|---|---|---|
| | | M12 N=22 | M24 N=21 | M36 N=19 | M48 N=18 | M72 N=16 | M96 N=5 |
| 6 | [reg] [naive] | .878 **.888** | .800 **.850** *** | .737 **.803** *** | .624 **.708** ** | .509 **.626** ** | .483 **.602** |
| 12 | [reg] [naive] | - - | .839 **.875** * | .769 **.832** *** | .658 **.735** ** | .523 **.644** ** | .465 **.608** * |
| 18 | [reg] [naive] | - - | .885 **.890** | .823 **.851** * | .738 **.764** | .611 **.661** | .579 **.627** |
| 24 | [reg] [naive] | - - | - - | .864 **.869** | .778 **.779** | .681 **.689** | **.657** .653 |
| max - 1 ~60 months | [reg] [naive] | **.807** .797 | Prediction at the most remote possible time point (~76 months) for all subjects (N=22). | | | | |

Table 1: Averaged Dice performance measures between predictions and observations for varying extents of learning datasets and extrapolation. The [reg] tag indicates the regression-based prediction, and [naive] the naive one. Each row corresponds to an increasingly large learning dataset, patients being observed for widening periods of time. Each column corresponds to an increasingly remote predicted visit from baseline. Significance levels [.05, .01, .001, .0001] for the Mann-Whitney test.

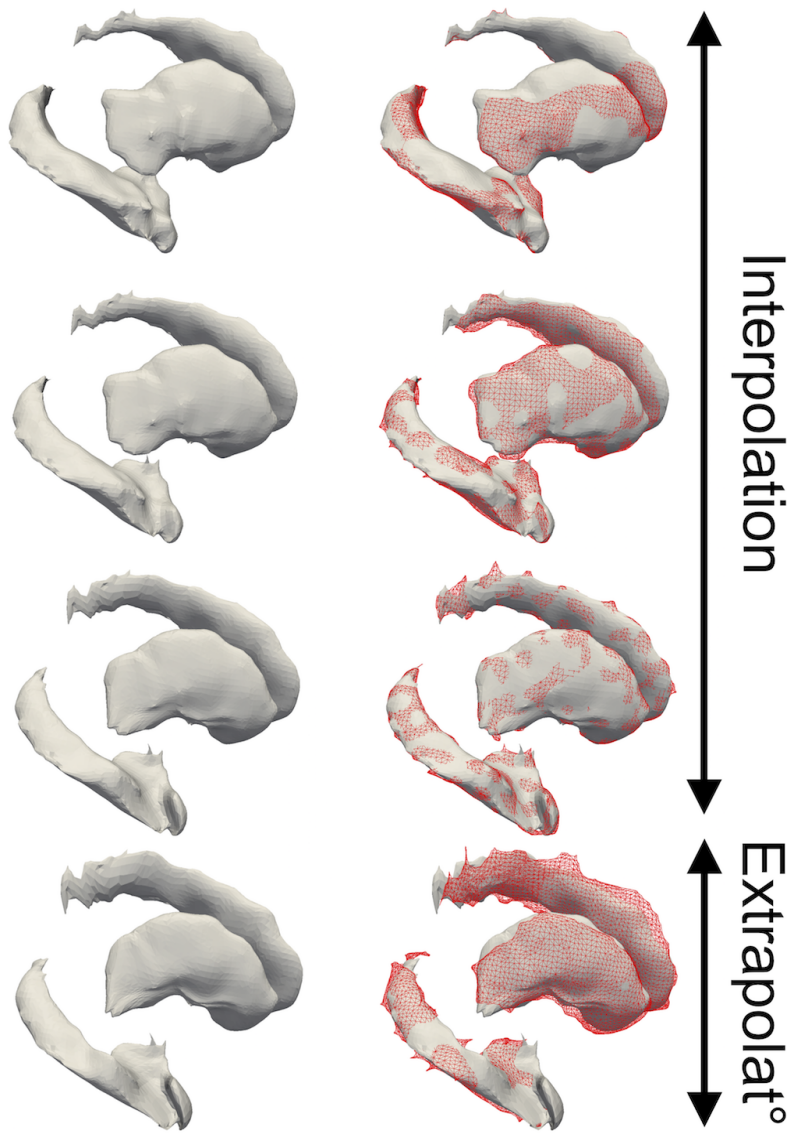

Fig. 1: Extrapolated geodesic regression for the subject s0671. Are only represented the right hippocampus, caudate and putamen brain structures in each subfigure. The three first rows present the interpolated brain structures, corresponding to ages 61.2, 64.2 and 67.2 (years). The last row presents the extrapolation result at age 70.2. On the right column are added the target brain structures (red wireframes), segmented from the original images.

Figure 1 displays an extrapolated geodesic regression for a specific subject, with a large learning period of 72 months, and a prediction at 108 months from the baseline (Dice performance of 0.74 versus 0.65 with the naive approach).

### 3.3 Non reparametrized transport

Among the 22 subjects whose regression-based predictive power has been evaluated in the previous section, the two which performed best are chosen as references for the rest of this paper. Their progressions are transported onto the 73 other subjects with the two different parallel shifting methods.

In more details, for each pair of reference and target subjects, the baseline target shape is first registered to the reference baseline. The reference geodesic regression is then either geodesically or exp-parallelized. Prediction performance is finally assessed : the Dice index between the prediction and the actual observation, for the two modes of transport, are computed and compared to the Dice index between the baseline meshes and the actual observation – the only available information in the absence of a predictive paradigm.

The upper part of Table 2 presents the results. In most cases, the obtained meshes by the proposed protocol are of lesser quality than the reference ones, according to the Dice performance metric. The two methods of transport are essentially similarly predictive, although geodesic parallelization slightly outperforms the exp-parallelization for the M12 prediction.

| Time reparam. | Method | Predicted follow-up visit | | | | | |
|---|---|---|---|---|---|---|---|
| | | M12 | M24 | M36 | M48 | M72 | M96 |
| | | N=144 | N=138 | N=130 | N=129 | N=76 | N=11 |
| Without reparam. | [exp] | .878 | .841 | .799 | .744 | .650 | .647 |
| | [geod] | **.883** | .847 | **.806** | .753 | .664 | **.661** |
| | [naive] | .882 | **.850** | .806 | **.754** | **.682** | .611 |
| | | N=140 | N=134 | N=123 | N=113 | N=62 | N=17 |
| With reparam. | [exp] | .882 | .852 | .825 | .796 | .756 | .730 |
| | [geod] | **.888** | **.858** | **.831** | **.802** | **.762** | **.732** |
| | [naive] | .884 | .852 | .809 | .764 | .706 | .636 |

Table 2: Averaged Dice performance measures between predictions and obervations for two modes of transport, with or without refinement by the cognitive scores. In each cell, the first line corresponds to the exp-parallelization-based prediction [exp], the middle line to the geodesic parallelization-based one [geod], and the last line to the naive approach [naive]. Each column corresponds to an increasingly remote predicted visit from baseline. Significance levels for the Mann-Whitney test [.05, .01, .001, .0001].

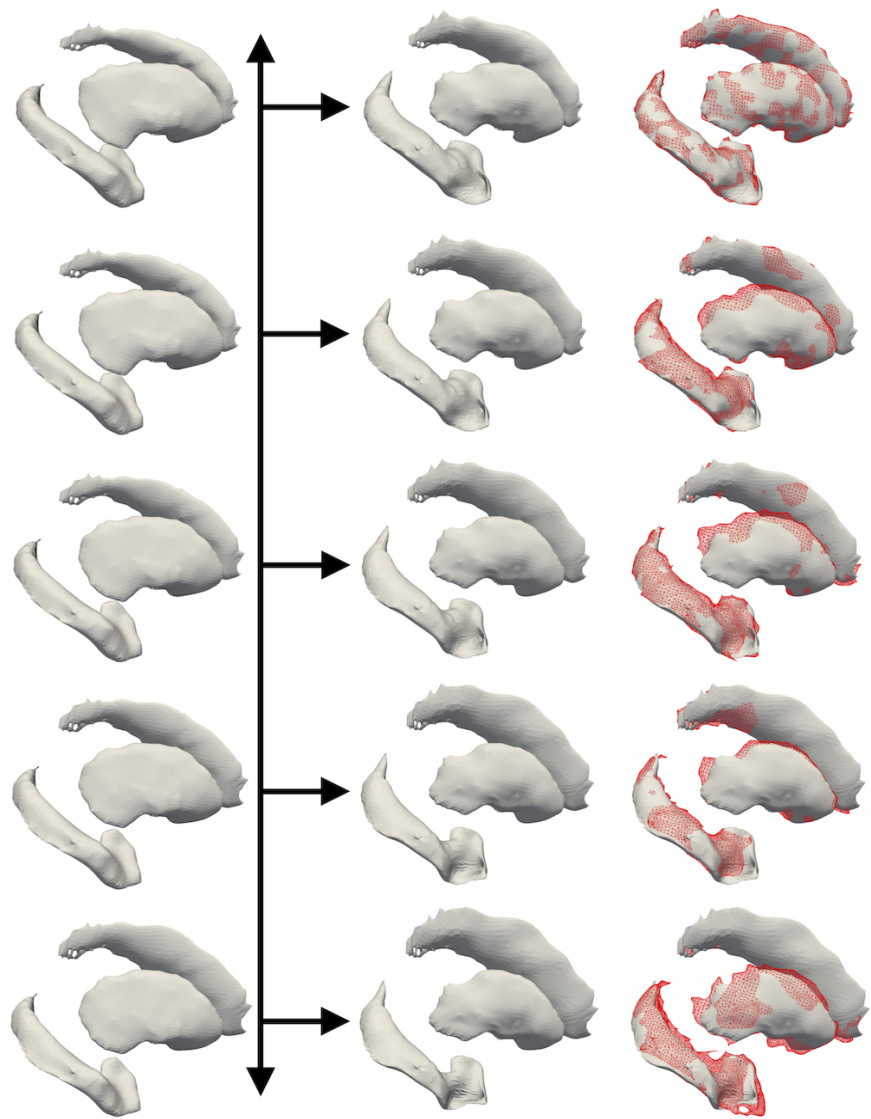

Fig. 2: Exp-parallelization of the reference subject s0906 (first column) towards the subject s1080 (second column), giving predictions for ages 81.6, 82.6, 83.6, 84.6 and 85.6 (years). On the third column are added the target brain structures (red wireframes), segmented from the original images.

### 3.4 Refining with cognitive dynamical parameters

The two reference progressions are transported through geodesic and exp-parallelization onto all remaining subjects. After time-reparametrization, the obtained parallel trajectories then deliver predictions for the brain structures.

Figure 3 displays a reference geodesic and an exp-parallelized curve. The predicted progression graphically matches the datapoints, and it can be noticed that the final prediction at age 85.6 (Dice 0.73) outperforms the corresponding one on Figure 2, obtained without time-reparametrization (Dice 0.69).

Quantitative results are presented in the lower part of Table 2. At the exception of the M12 prediction, both protocols outperform the naive one. The M36, M48, M72 and M96 predictions are the most impressive ones, with p-values always lesser than 1%. This shows that the pace of cognitive score evolution is well correlated with the pace of structural brain changes, and therefore allows an enhanced prediction of follow-up shapes.

No conclusion can be drawn concerning the two parallel shifting methodologies, a single weak significance result being obtained only for the M12 prediction where the geodesic parallelization method slightly outperforms the exp-parallelization one with a Dice score of 0.888 versus 0.882.

## 4 Conclusion

We conducted a quantitative study of geodesic regression extrapolation, exhibiting its limited predictive abilities. We then proposed a method to transport a spatiotemporal trajectory into a different subject space with cognitive decline-derived time reparametrization, and demonstrated its potential for prognosis. The results show how crucial the dynamics are in disease modeling, and how cross-modality data can be exploited to improve a learning algorithm. The two main paradigms that have emerged for the transport of parallel trajectories were shown to perform equally well in this prediction task. Nonetheless, the exp-parallelization offers a methodological advantage in that the generated trajectories do not depend on a particular choice of point on the reference geodesic, in contrast with the trajectories obtained by geodesic parallelization. It takes full advantage of the isometric property of the parallel transport, and eases the combination with time-warp functions based on the individual disease dynamics.

In future work, more complex time reparametrization could be considered as in [7]. Finally, the robustness of the proposed protocol to the choice of reference subject has not been assessed. Such a choice could be avoided by constructing an average disease model as in [15], or by translating for shapes the method of [14]. We may also use this framework to estimate a joint image and cognitive model to better estimate individual dynamical parameters of disease progression.

**Acknowledgments.** This work has been partly funded by the European Research Council (ERC) under grant agreement No 678304, European Union's Horizon 2020 research and innovation program under grant agreement No 666992, and the program Investissements d'avenir ANR-10-IAIHU-06.

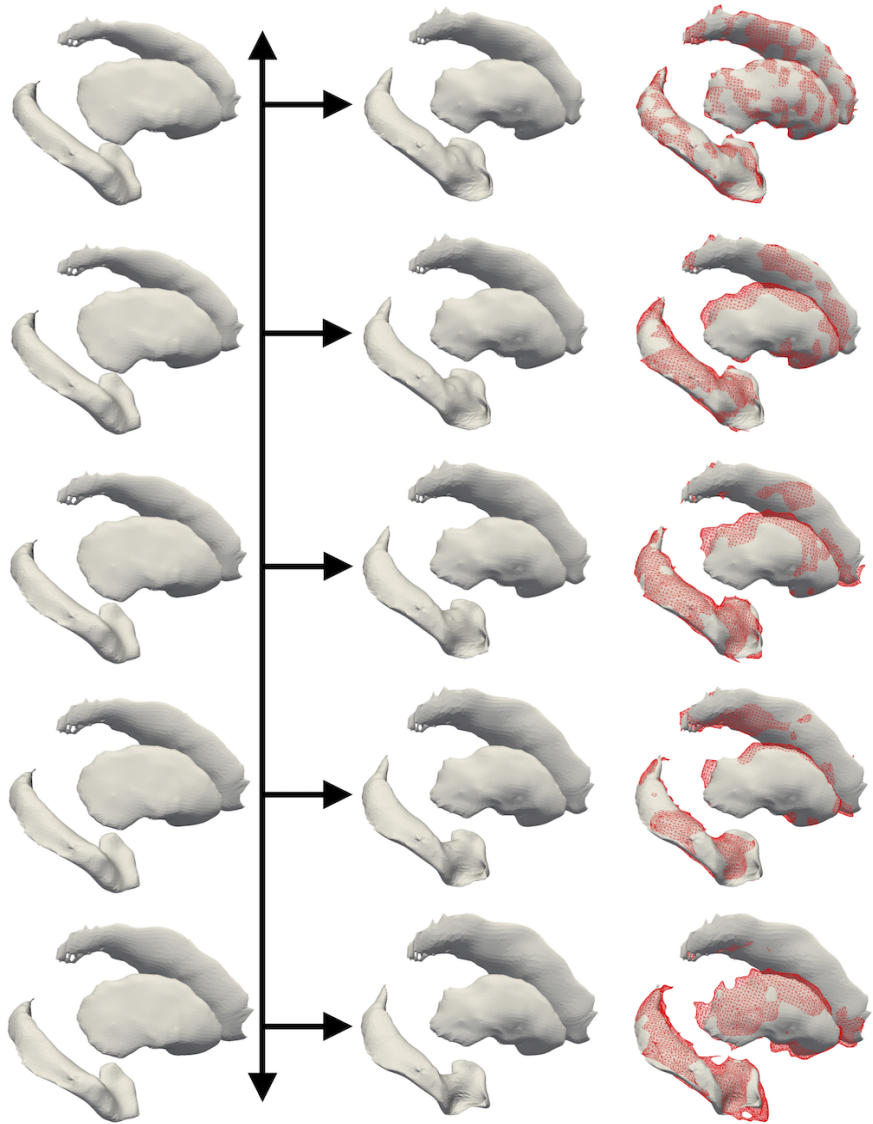

Fig. 3: Time-reparametrized exp-parallelization of the reference subject s0906 (first column) towards the subject s1080 (second column), giving predictions for ages 81.6, 82.6, 83.6, 84.6 and 85.6 (years). On the third column are added the target brain structures (red wireframes), segmented from the original images.

# References

1. Beg, M., Miller, M., Trouvé, A., Younes, L.: Computing large deformation metric mappings via geodesic flows of diffeomorphisms. IJCV 61(2), 139–157 (2005)
2. Durrleman, S., Allassonnière, S., Joshi, S.: Sparse adaptive parameterization of variability in image ensembles. IJCV 101(1), 161–183 (2013)
3. Durrleman, S., Prastawa, M., Charon, N., Korenberg, J.R., Joshi, S., Gerig, G., Trouvé, A.: Morphometry of anatomical shape complexes with dense deformations and sparse parameters. NeuroImage (2014)
4. Fishbaugh, J., Prastawa, M., Gerig, G., Durrleman, S.: Geodesic regression of image and shape data for improved modeling of 4D trajectories
5. Fletcher, T.: Geodesic regression and the theory of least squares on riemannian manifolds. IJCV 105(2), 171–185 (2013)
6. Hadj-Hamou, M., Lorenzi, M., Ayache, N., Pennec, X.: Longitudinal analysis of image time series with diffeomorphic deformations: A computational framework based on stationary velocity fields. Frontiers in Neuroscience 10, 236 (2016)
7. Hong, Y., Singh, N., Kwitt, R., Niethammer, M.: Time-warped geodesic regression. In: MICCAI. vol. 17, p. 105 (2014)
8. Lorenzi, M., Ayache, N., Frisoni, G., Pennec, X.: 4D registration of serial brains MR images: a robust measure of changes applied to Alzheimer's disease. Spatio Temporal Image Analysis Workshop (STIA), MICCAI (2010)
9. Lorenzi, M., Ayache, N., Pennec, X.: Schild's ladder for the parallel transport of deformations in time series of images. pp. 463–474. Springer (2011)
10. Lorenzi, M., Pennec, X.: Geodesics, parallel transport & one-parameter subgroups for diffeomorphic image registration. IJCV 105(2), 111–127 (Nov 2013)
11. Metz, C., Klein, S., Schaap, M., van Walsum, T., Niessen, W.: Nonrigid registration of dynamic medical imaging data using nd + t b-splines and a groupwise optimization approach. Medical Image Analysis 15(2), 238 – 249 (2011)
12. Peyrat, J., Delingette, H., Sermesant, M., Pennec, X., Xu, C., Ayache, N.: Registration of 4D time-series of cardiac images with multichannel diffeomorphic Demons. MICCAI (2008)
13. Qiu, A., Younes, L., Miller, M.I., Csernansky, J.G.: Parallel transport in diffeomorphisms distinguishes the time-dependent pattern of hippocampal surface deformation due to healthy aging and the dementia of the Alzheimer's type. NeuroImage 40(1), 68–76 (2008)
14. Schiratti, J.B., Allassonnière, S., Colliot, O., Durrleman, S.: Learning spatiotemporal trajectories from manifold-valued longitudinal data. In: NIPS 28
15. Singh, N., Hinkle, J., Joshi, S., Fletcher, P.T.: Hierarchical geodesic models in diffeomorphisms. IJCV 117(1), 70–92 (2016)
16. Vercauteren, T., Pennec, X., Perchant, A., Ayache, N.: Non-parametric diffeomorphic image registration with the Demons algorithm. In: MICCAI
17. Wang, L., Beg, F., Ratnanather, T., Ceritoglu, C., Younes, L., Morris, J.C., Csernansky, J.G., Miller, M.I.: Large deformation diffeomorphism and momentum based hippocampal shape discrimination in dementia of the Alzheimer type. IEEE Transactions on Medical Imaging 26(4), 462–470 (2007)
18. Wu, G., Wang, Q., Lian, J., Shen, D.: Estimating the 4D respiratory lung motion by spatiotemporal registration and building super-resolution image. In: MICCAI. pp. 532–539 (2011)
19. Younes, L.: Jacobi fields in groups of diffeomorphisms and applications. Quarterly of Applied Mathematics 65(1), 113–134 (2007)